\pdfoutput=1

\documentclass[11pt]{article}

\usepackage{acl}

\usepackage{times}
\usepackage{latexsym}

\usepackage[T1]{fontenc}

\usepackage[utf8]{inputenc}

\usepackage{microtype}
\usepackage{tikz}
\usepackage{comment}
\usepackage{amsmath,amssymb} 
\usepackage{color}
\usepackage{cite}
\usepackage{booktabs}
\usepackage{epsfig}
\usepackage{graphicx}
\usepackage{caption}
\usepackage{subfigure}
\usepackage{todonotes}

\usepackage{multirow}
\usepackage{verbatim}
\usepackage{bm}

\newcommand*{\ie}{\textit{i.e.}}
\newcommand*{\eg}{\textit{e.g.}}
\newcommand*{\modelname}{FineCo}

\title{Contrastive Video-Language Learning with Fine-grained Frame Sampling}

\author{Zixu Wang$^{1}$, Yujie Zhong$^{2}$, Yishu Miao$^{3}$, Lin Ma$^{2}$, Lucia Specia$^{1}$ \\
  $^{1}$Language and Multimodal AI Lab (LAMA), Imperial College London \\
  $^{2}$Meituan Inc., $^{3}$Haiper.ai \\
  \texttt{\small zixu.wang@imperial.ac.uk, jaszhong@hotmail.com, yishu.miao@haiper.ai} \\
  \texttt{\small forest.linma@gmail.com, l.specia@imperial.ac.uk} \\
  }

\begin{document}
\maketitle
\begin{abstract}
Despite recent progress in video and language representation learning, the weak or sparse correspondence between the two modalities remains a bottleneck in the area.
Most video-language models are trained via pair-level loss to predict whether a pair of video and text is aligned.
However, even in paired video-text segments, only a subset of the frames are semantically relevant to the corresponding text, with the remainder representing noise; where the ratio of noisy frames is higher for longer videos.
We propose \textbf{\modelname} (\underline{\textbf{Fine}}-grained \underline{\textbf{Co}}ntrastive Loss for Frame Sampling), an approach to better learn video and language representations with a fine-grained contrastive objective operating on video frames. 
It helps distil a video by selecting the frames that are semantically equivalent to the text, improving cross-modal correspondence.
Building on the well established VideoCLIP model as a starting point, \modelname\ achieves state-of-the-art performance on YouCookII, a text-video retrieval benchmark with long videos.
\modelname\ also achieves competitive results on text-video retrieval  (MSR-VTT), and video question answering datasets (MSR-VTT QA and MSR-VTT MC) with shorter videos.
\end{abstract}

\section{Introduction}
Human perception is multimodal, including visual, textual, and audial information.
To achieve human-level perceptional ability, intelligent systems need to understand and interpret these multimodal signals and summarise the relevant information in them.
Learning from video and language data has received significant attention in recent multimodal machine learning work for downstream tasks that require joint understanding of video and textual information, 
including text-video retrieval \citep{10.1007/978-3-319-10602-1_48, Liu2019a, Miech2018LearningAT, Wang_2016_CVPR, Bain21}, 
video question answering \citep{fan-CVPR-2019,Yang_2021_ICCV, L-GCN2020AAAI, Jiang_Chen_Lin_Zhao_Gao_2020, le2020hierarchical, lei2021less}, 
and video captioning \citep{ging2020coot, Luo2020UniVL, Zhang_2020_CVPR}.
In most of this work, contrastive learning \citep{pmlr-v9-gutmann10a} is used as training objective. 

The aim of a cross-modal contrastive loss is to maximise the similarity between an aligned video-text pair while minimising the similarity for all other pairs.
One issue with standard cross-modal contrastive loss is that it focuses on pair-level alignment but ignores the negative effects of irrelevant frames that are present in a single video clip, even in a pair of aligned video and text.
We define irrelevant frames as those with no or little shared semantics with the text.
These irrelevant frames may negatively affect the contribution of frames that are semantically similar to the text, which further results in less informative video representation.
Therefore, we posit that frame-level learning is a better strategy for video-language tasks. 

\begin{figure}[t]
\centering
\includegraphics[width=0.85\linewidth]{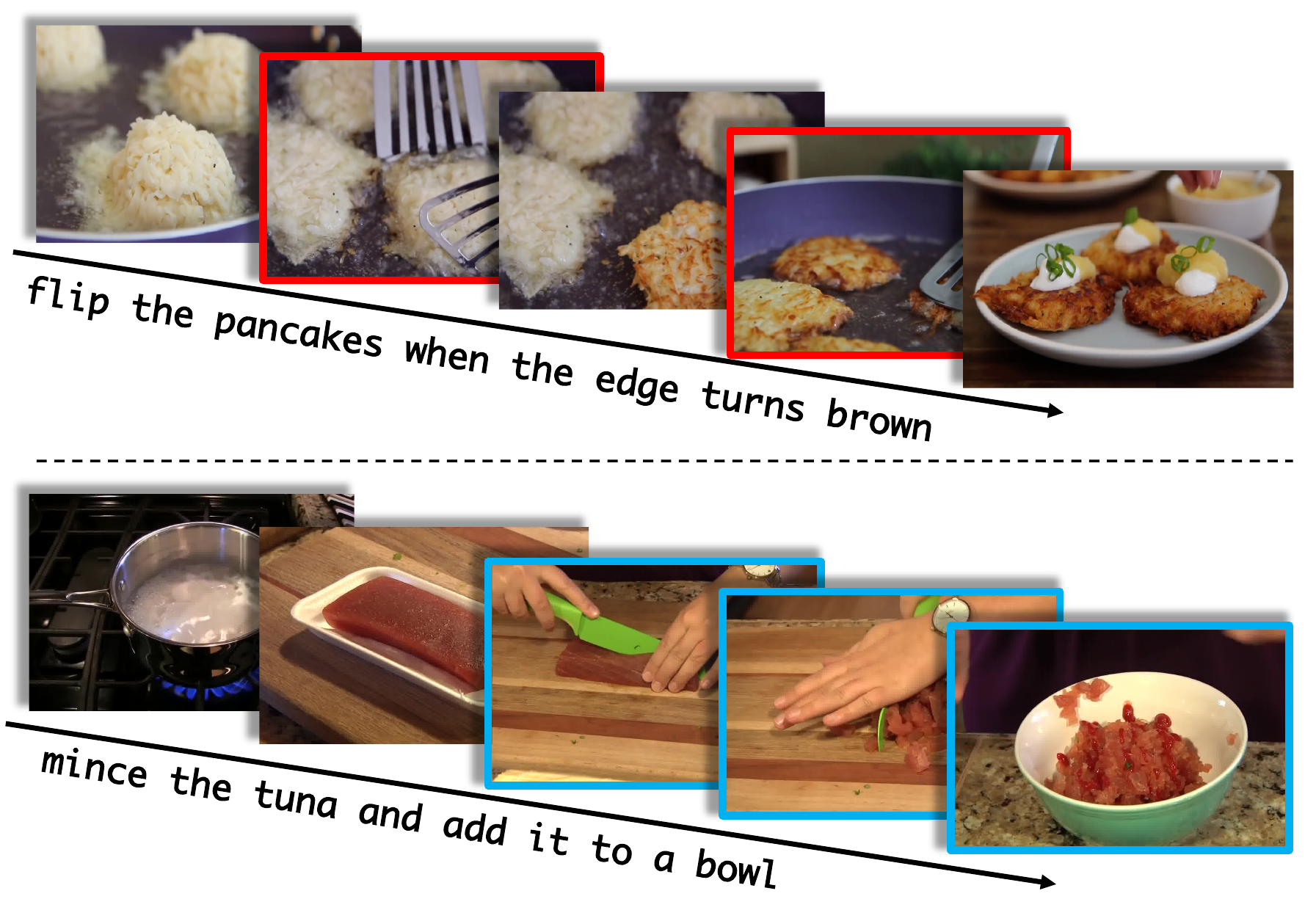}
\caption{Illustration of the weak correspondence problem in video-language learning. Given a pair of video and its text (\eg\ caption, instruction, or transcription), only a subset of the frames (here indicated by coloured bounding boxes) is semantically aligned to the textual content. The remaining frames represent irrelevant visual information and will not contribute to language grounding on videos.}
\label{fig:illustration}
\end{figure}

In this paper, we propose \modelname, an approach that has  a frame selector to sample relevant frames in a video and is trained with a fine-grained contrastive loss on frame-text pairs, in order to mitigate the problem of weak correspondence in video-language representation learning.
Existing video-language learning approaches \citep{miech19endtoend, xu-etal-2021-videoclip} only optimise pair-level alignment but do not explicitly learn which part of a video contributes to its alignment with the text. 
\modelname\ focuses on aligning relevant frames with the text. It is inspired by the text-based temporal localisation task \citep{NEURIPS2020_d27b95ca}, 
however, the motivation of \modelname\ is different: to learn better video-level representation by adding a frame-level contrastive learning signal to the pair-level objective, with no need for temporal annotation within a video-text pair.

We hypothesise that \modelname\ is particularly beneficial for long videos, where each video provides more information and only a small proportion of frames will be relevant to its text counterpart, as shown in Figure \ref{fig:illustration}.
\modelname\ is able to model frame-text similarity through fine-grained contrastive learning, where the most informative frames are paired with the text as positive pairs and the remaining frames, as negatives.
It then explicitly contrasts the selected informative frames against the noisy frames, without the need for frame-text annotations.
This frame-level distillation provides a strong learning signal, which encourages the alignment of semantically equivalent video-text pairs.
The fine-grained contrastive loss abstracts the learning signal from pair-level annotations and is trained in an end-to-end manner.
This combination of pair-level learning signal and frame-level contrastive loss is novel and effective, and boosts the performance on two important video-language benchmark tasks, especially in text-video retrieval with longer videos.
We devised \modelname\ by building on the recently proposed and well performing VideoCLIP \citep{xu-etal-2021-videoclip}, in which a video clip is represented as sequence of frame features. 

Our contributions are summarised as follows:
(1) We propose \modelname, an approach trained with fine-grained contrastive loss to mitigate the weak correspondence problem in video-text pairs;
(2) We use \modelname\ to distil a video clip by sampling  frames that are relevant to its  text counterpart according to frame-text similarities;
(3) On text-video retrieval and video question answering benchmarks, we show that  \modelname\ achieves 
state-of-the-art performance on YouCookII and MSR-VTT MC (multiple choice).

\section{Related Work}
\paragraph{Contrastive Learning}
The use of contrastive loss \citep{pmlr-v9-gutmann10a} has become the dominant paradigm for learning video-language representations.
The aim is to maximise the similarity of  video-text pairs that are aligned to each other (positive pairs) while pushing away irrelevant (negative) pairs.
However, the semantic alignment between most video-text pairs is weak, which makes it difficult to ground textual information on the videos.
In order to mitigate the pair-level weak alignment issue, MIL-NCE \citep{miech19endtoend} leverages multiple surrounding captions as the positive pairs and makes use of multiple instance learning (MIL) \citep{DIETTERICH199731} with contrastive loss to mitigate noise in cross-modal correspondences.
The main idea is to consider multiple contextual sentences for matching a video, instead of only comparing a video against a single sentence.
To alleviate the issue that semantically equivalent videos and texts from different pairs may be taken as dissimilar in contrastive learning,  
support-set \citep{patrick2021supportset} introduces a generative approach for captioning over a set of visual candidates that ensures that video-language representation does not over specialise to individual samples.
MIL-NCE and support-set focus on pair-level contrastive signals to align relevant video-text pairs. 
However, even within a positive video-text pair, the video is likely to contain many irrelevant frames. Therefore, it can be beneficial to distil the video such that only the relevant frames, \ie\ those which have similar content to the text, are selected for cross-modal learning.

\paragraph{Video-language Learning} \citep{Sun_2019_ICCV, Zhu_2020_CVPR, gabeur2020mmt, li2020hero, miech19endtoend, ging2020coot, Luo2020UniVL}
have shown promising results for video-language learning with pre-training followed by fine-tuning.
This strategy has become very prominent since the release of BERT \citep{devlin-etal-2019-bert} and many image-text pre-training frameworks \citep{tan-bansal-2019-lxmert, li2019visualbert, li2020oscar, zhang2021vinvl, chen2020uniter, zhang-etal-2019-ernie, pmlr-v139-kim21k, ALBEF, li2022blip}.
The release of datasets such as HowTo100M \citep{miech19howto100m} and WebVid-2M \citep{Bain21} has enabled large-scale pre-training on unlabelled video-text pairs to improve representation learning of video and language.
Many approaches \citep{miech19endtoend, Zhu_2020_CVPR, patrick2021supportset} use HowTo100M as their pre-training dataset. 
FiT \citep{Bain21} uses WebVid-2M and Google Conceptual Captions (CC3M) to take advantage of the large collection of video-text and image-text pairs for pre-training.
However, large pre-training datasets rely on loosely aligned video-text pairs, without any fine-grained supervision on alignment.
This makes it difficult to learn cross-modal cues present in the given video-text pairs. 
It is also computationally expensive to improve video-language representation learning, given that videos can contain a large number of frames, especially longer videos.
ClipBERT \citep{lei2021less} randomly samples a few frames from a video for video-language representation learning.
Their motivation is to minimise memory and computation costs from processing the full sequence of frames. 
This sampling strategy is over simplistic and can thus be improved by better approaches to select frames based on their relevance to the paired text.


\section{\modelname}

\subsection{Preliminaries}
The most widely used objective function for video-language learning is contrastive loss, specifically the softmax version of noise-contrastive estimation (NCE) \citep{pmlr-v9-gutmann10a}. It is formulated as
\begin{equation}
\label{eq:contra}
    \sum^{n}_{i=1}\log\left(
    \frac{e^{f(x_{i})^{T}g(y_{i})}}
    {e^{f(x_{i})^{T}g(y_{i})}+
    \sum\limits_{(x', y')\in \mathcal{N}_{i}}e^{f(x'_{i})^{T}g(y'_{i})}} \right)
\end{equation}
where $x_{i}$ denotes a video clip and $y_{i}$ represents the corresponding text (\eg\ a caption, an instruction, or transcription);
$f$ and $g$ are video encoder and text encoder respectively;
$e^{f(x_{i})^{T}g(y_{i})}$ denotes the similarity of a positive video-text pair, calculated as the exponentiated dot product of the video representation $f(x_{i})$ and text representation $g(y_{i})$;
$\mathcal{N}_{i}$ is a set of negative video-text pairs $x'_{i}$ and $y'_{i}$ that are not aligned.

This contrastive loss leverages pair-level similarity of video and text, but ignores the fact that weak video-language correspondence does not stem only from entirely negative pairs of video and text, but 
also from frame-level noise, which happens even when a video-text pair is aligned as a whole.
Standard contrastive loss does not explicitly model frame-text relevance, \ie\ it does not differentiate between frames that are semantically equivalent to the corresponding text and frames that are not. 
It can thus suffer by learning from noisy signals, particularly in long videos with various scenes. 

\begin{figure*}[t]
\centering
\includegraphics[width=0.8\linewidth]{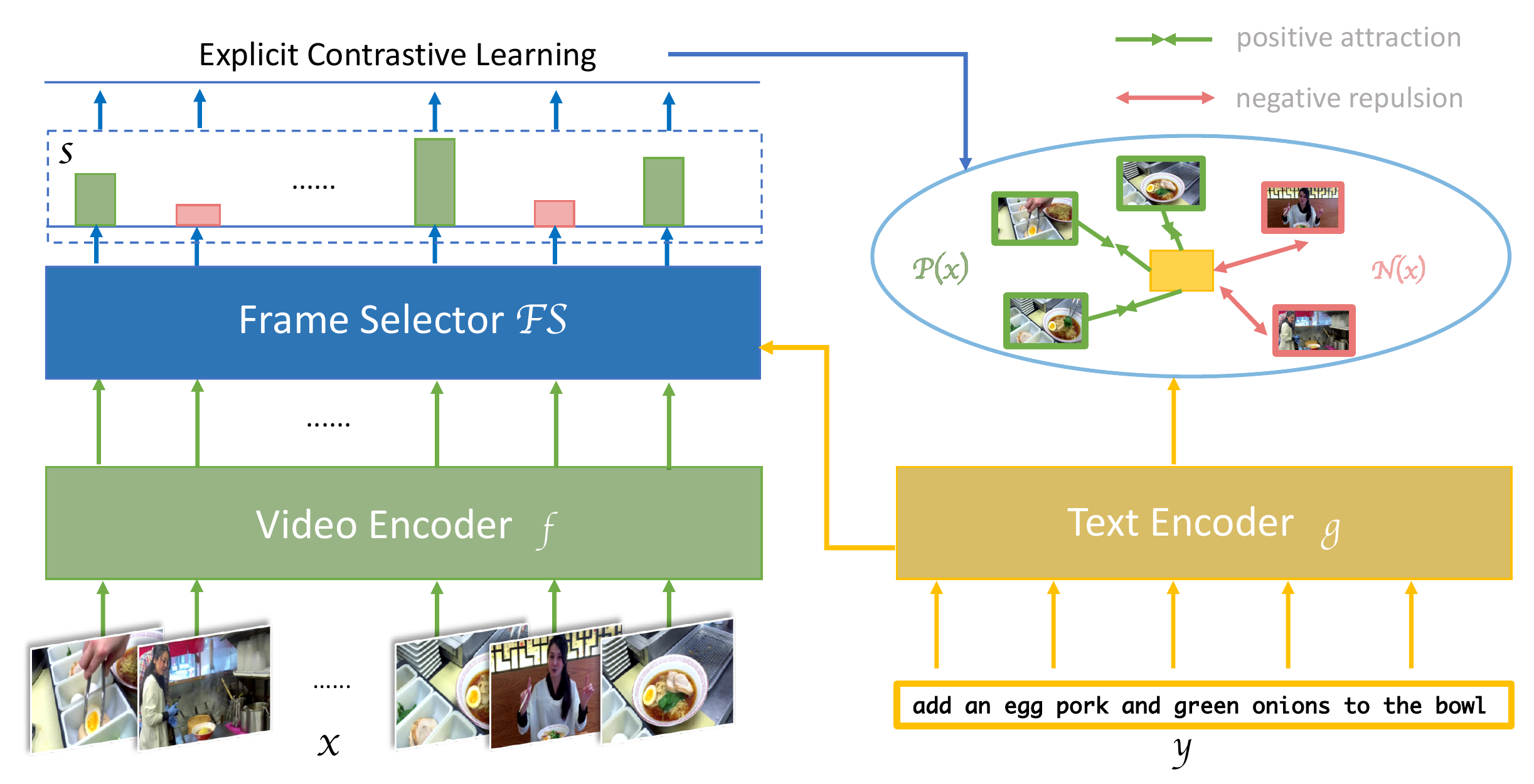}
\caption{\modelname\ architecture. 
Given a sequence of frames in a video clip $x$, the video encoder $f$ transforms them into a sequence of video features.
The corresponding sentence $y$ is fed into the text encoder $g$ to get the text representation.
The frame selector $FS$ takes the text representation and the sequence of video features as inputs and outputs the similarities (probabilities of each frame being relevant).
The top $k$ frames are then used as the positive candidates and the remaining ones as negative, both of which are combined with the text representation to compute the fine-grained contrastive loss.}
\label{fig:model}
\end{figure*}

\subsection{Fine-grained Contrastive Learning}
A video consists of a sequence of frames.
For video-language learning, the video is paired with a text which describes/refers to some of the content of the video. 
For most tasks, only some of the visual information has an equivalent textual signal, \eg\ a video description is only a summary of the visual information.
To sample and optimise for the relevant visual information from a video, we propose a fine-grained contrastive loss to distil each video-text pair.

Formally, a video-text pair is denoted as ($x$, $y$), 
where $x$ is a video clip consisting of a sequence of $N$ video frames $\{x_1, x_2, \dots, x_K\}$ where $K$ is the number of frames in the video clip, and $y$ is the paired text.
We assume that a video $x$ contains a set of $C$ positive frames $\mathcal{P}(x)$ and a set of $(K-C)$ negative frames $\mathcal{N}(x)$, 
where positive frames contains relevant information to the text while negative frames are noisy/irrelevant ones.
The aim is to maximise the joint probability of relevant frame-text pairs $(x_k, y)$ by exponentiating the similarity of the two representations:
\begin{equation}
    p(x_k,y) = h(f(x_k), g(y)) \propto e^{\text{sim}(f(x_k), g(y))}
\end{equation}

\subsubsection{Objective Function} 
Given $n$ pairs of video representation $f(x)$ and text representation $g(y)$,
the $i$th pair is denoted as $f(x_i) = \{f(x_{i_1}), f(x_{i_2}), \dots, f(x_{i_K})\}$ and $g(y_i)$,
our fine-grained contrastive loss $\mathcal{L}$ is defined as: 
\begin{equation}
    \begin{aligned}
        \mathcal{A}_i &= \sum\limits_{x_{i_k}\in \mathcal{P}(x_i)}e^{\text{sim}(f(x_{i_k}), g(y_{i}))} \\ 
        \mathcal{B}_i &= \sum\limits_{x'_{i_k}\in \mathcal{N}(x_i)}e^{\text{sim}(f(x'_{i_k}), g(y_{i}))} \\
        \mathcal{L} &= \sum^{n}_{i=1}\log\left(
        \frac{\mathcal{A}_i}{\mathcal{A}_i + \mathcal{B}_i} \right)
    \end{aligned}
\end{equation}
where $\mathcal{P}(x_i)$ contains the positive frames in a video that have higher similarities to the text representation $g(y_i)$, 
and $\mathcal{N}(x_i)$ is the set of remaining frames in the same video, which refers to the negative frames.
The similarity is calculated by our frame selector ($\mathcal{FS}$) (Section \ref{sec:fs}) with the frame $x_{i_k}$ and text representations $y_{i}$ as inputs.
$\mathcal{A}_i$ and $\mathcal{B}_i$ represent the sum of similarity scores for positive and negative frames, respectively. 
This objective function aims to maximise the similarity between the positive frames and the text, while increasing the dissimilarity between the negative frames and the text.
Therefore, the sampled relevant frames can directly contribute to the cross-modal learning of video-text alignments.

\subsubsection{Assignment of Positives and Negatives}
\label{sec:fs}
Inspired by MIL-NCE \citep{miech19endtoend}, which makes use of multiple sentences for matching a video and its corresponding text, 
we extract multiple positive frames from the complete set according to the similarity score between each frame and the text.
Consider an example $(x, y)$ with $K$ frames $\{x_1, x_2, \dots, x_K\}$, 
we introduce a frame selector $\mathcal{FS}$, a cross-modal module which takes video and text representation as the input and outputs the similarity scores between each frame and the text, denoted as:
\begin{equation}
    \text{sim}_k = \mathcal{FS}(f(x_k), g(y)); x_k \in \{x_1, x_2, \dots, x_K\}
\end{equation}
where $f(x_k)$ is the representation of the $k$th frame;
$g(y)$ is the representation that encodes the meaning of the complete text sequence, which is used to find semantically similar frames in the corresponding video $x$;
$\text{sim}_k$ is the similarity score between the $k$th frame and the text $y$.

By ranking the similarity scores of $K$ frames, we choose top $C$ frames to form the positive set and the remaining $(K-C)$ as the negative set.
This is an explicit sampling strategy which extracts the relevant frames in a video.
There is no constraint on the architecture of $\mathcal{FS}$.
In this work, we use a multi-layer perceptron (MLP) with a softmax layer to compute the similarity scores.




\subsection{Model Architecture}
As our methodology focuses on fine-grained contrastive learning signal for a single pair of video and its text,
it makes no assumptions on the encoder architectures and can work with pre-training frameworks with different video and text backbones.
In our experiments, we use Transformer \citep{NIPS2017_3f5ee243} as both the video encoder and the text encoder, as we detail below.

\subsubsection{Text Encoder}
We use BERT \citep{devlin-etal-2019-bert} as the text encoder $g$ to get text representation $g(y)$.
The text encoder is trained together with the video encoder to learn better text representations.
Following VideoCLIP \citep{xu-etal-2021-videoclip}, we use average pooling (instead of using the \texttt{[CLS]} token) as the final text encoding.
The text representation is used as the guiding element and anchor to calculate the frame-text similarity scores and to sample the most semantically similar frames in a video clip.

\subsubsection{Video Encoder}
Our video encoder $f$ is composed of an S3D \citep{Xie_2018_ECCV, miech19endtoend} and a Transformer \citep{NIPS2017_3f5ee243}, following VideoCLIP \citep{xu-etal-2021-videoclip}.
To speed up training, we use a S3D pre-trained on HowTo100M \citep{miech19howto100m} to extract pre-trained video features, where the video feature of a video clip is represented by a sequence of video frames.
The output from the S3D is formulated as $x = [x_{1}, x_{2}, \dots, x_{K}]$, where $x$ is the representation of a sequence of video frames. 
We extract the frames at a rate of one frame per second, so the number of video frames equals the number of seconds.
$x$ is concatenated with learnable tokens \texttt{[CLS]} and \texttt{[SEP]} at the beginning and the end of the sequence, respectively.
We then train the Transformer using the pre-extracted video representation as the input, to obtain the last hidden states as the representation of the sequence of video frames.


\subsection{Training}
Training with the pair-level contrastive loss is challenging due to the intractability of computing the normalisation constant over all possible pairs of videos and texts.
It is however more feasible in our fine-grained contrastive loss as the number of possible frames in a single video clip is limited. 
The normalising constant is computationally tractable and can be directly computed by summing over exponentiated similarity scores across all the frame-text pairs.
The overall training objective ($\mathcal{L}$) is defined by combining our fine-grained contrastive loss ($\mathcal{L}_1$) and task-specific losses ($\mathcal{L}_2$), denoted by $\mathcal{L} = \mathcal{L}_1 + \mathcal{L}_2$;
where in text-video retrieval, the task loss $\mathcal{L}_2$ is pair-level contrastive loss and in video question answering, it is cross-entropy.

\subsection{Inference}
For text-video retrieval, there is no cross-modal fusion module at inference time.
It requires only video and text representations which are first projected to a common dimension via linear layers. The similarity between a video-text pair is calculated by performing the exponentiated dot product between the two projected embeddings.
This ensures retrieval inference is of trivial cost, since it is indexable and scalable to large-scale retrieval at inference time.
For video question answering, we follow the pipeline in Figure~\ref{fig:model}, where we concatenate the video and text representations, and feed it into an MLP module to obtain the final representation for answer prediction.

\section{Experiments}
In this section, we describe the tasks and datasets used in our experiments with \modelname.

\subsection{Datasets and Metrics}
\modelname\ is mainly beneficial for long videos, therefore we focus our evaluation on YouCookII \citep{ZhXuCoCVPR18} - a text-video retrieval dataset with long videos.
\textbf{YouCookII}
consists of 2K cooking videos with 14K video clips. 
The videos are of a total duration of 176 hours with average \textbf{5.26 minutes} per video. 
Each video clip is annotated with one sentence on a cooking instruction.
It is collected from YouTube and contains 89 types of recipes. 
We split the dataset according to \citet{miech19endtoend} where 9.6k video-text pairs are used for training and 3.3k pairs for validation.

We further evaluate \modelname\ on other benchmark datasets for text-video retrieval and video question answering with shorter videos. 
\textbf{MSR-VTT} \citep{xu2016msr-vtt}
is another popular benchmark dataset for text-video retrieval.
It contains 10K YouTube videos (an average \textbf{20 seconds} per video) with 200K captions.
We report the results on the \textbf{1k} test split and use the remaining 9k
videos for training.
\textbf{MSVD} \citep{chen-dolan-2011-collecting}
consists of 80K captions for 1,970 videos from YouTube, with each video containing 40 sentences.
We use the standard split of 1200, 100, and 670 videos for training, validation, and testing as in \citep{Liu2019a, patrick2021supportset}.
\textbf{DiDeMo} \citep{hendricks18emnlp}
contains 10K Flickr videos with 40K sentences. 
Following \citep{Liu2019a, lei2021less}, we evaluate paragraph-to-video retrieval, where all sentence descriptions from a video are concatenated into a single query.
\textbf{MSR-VTT QA} contains 10K videos and 243K open-ended questions, which is created using the videos and captions from original MSR-VTT.
We use 1500 most frequent answers as the answer vocabulary, which covers over 93\% samples.
\textbf{MSR-VTT MC} (multiple choice) is also created from original MSR-VTT.
Multiple choice QA is formulated as a video-text retrieval task where the videos are the questions and captions are the answers.

\paragraph{Evaluation Metrics}
Following the standard evaluation protocols as described in most video-language work \citep{miech19howto100m, DBLP:conf/eccv/ZhangHS18, 10.1145/3206025.3206064, Miech2018LearningAT, miech19endtoend},
we report the text-video retrieval performance using recall-based metrics: Recall at rank K (R@K) which measures the rate at which the correct video is retrieved amongst the top ranked results, and Median Rank (MdR) which calculates the median of a list of indices representing the rank of the ground truth video; where the higher R@K and lower median rank indicate better performance.
For MSR-VTT QA and MSR-VTT MC, accuracy is reported, as in \citet{xu-etal-2021-videoclip}.

\subsection{Training Details}
To minimise computation costs, we use
S3D \citep{Xie_2018_ECCV} for video feature extraction, which is pre-trained on HowTo100M \citep{miech19howto100m} following MIL-NCE \citep{miech19endtoend}.
The feature dimensionality is 512 (\eg\ given a 10-second video, the shape of the video feature extracted  is $[10, 512]$).
We apply video feature pre-extraction to all the downstream datasets in our experiments.
We follow the pre-training steps as in VideoCLIP \citep{xu-etal-2021-videoclip} where pre-training is done using HowTo100M, which contains uncurated instructional videos. A total of 1.1M videos are used for pre-training after cleaning and filtering.

For the video Transformer encoder, we use 6 attention blocks, while for the text Transformer encoder, we use 12 blocks.
The weights for both encoders are initialised with \textit{bert-base-uncased}.
The maximum length of a video is 32; for text inputs it is 64.
Before feeding video and text inputs into their respective encoders, [CLS] and [SEP] tokens are concatenated to the beginning and end of each modality.
All the models are trained on one NVIDIA Tesla V100 GPU with 32 GB of RAM memory for 15 epochs, with fp16 precision for 2-3 hours.
We select the final checkpoint according to the loss on the validation set.
Optimisation is performed using Adam \citep{DBLP:journals/corr/KingmaB14} with a learning rate of 5e-5.
The model takes 1000 steps for warm-up, and we use a learning rate schedule with polynomial decay.

\section{Results}
In this section, we describe the experimental results and compare \modelname\ with state-of-the-art approaches (Section \ref{subsec:results}).
We further explore different sampling strategies to select positive frames (Section \ref{subsec:number}), and fine-grained word sampling (Section \ref{subsec:word}).
We also provide examples of the frames selected by \modelname\ (Section \ref{subsec:examples}).

\subsection{Comparison to State-of-the-art}
\label{subsec:results}
Overall, as we detail below, \modelname\ outperforms its base model VideoCLIP across all benchmark datasets. 
Additionally, it achieves state-of-the-art performance on YouCookII and MSR-VTT MC.

\subsubsection{Text-video Retrieval}
We start by evaluating on YoucookII, which contains longer videos than other text-video benchmarks, and is therefore more challenging for video-language representation learning.
As shown in Table \ref{tab:youcook},
\modelname\ outperforms all previous approaches by a large margin.
We report results w/ and w/o Dual Softmax (DS) following \citet{cheng2021improving} and \citet{gao2021clip2tv}.
In Dual Softmax, given a similarity matrix in text-video retrieval, a prior probability is calculated in the cross direction, which is then multiplied with the original similarity matrix as an efficient regulariser.
\modelname\ surpasses previous state-of-the-art with fine-grained contrastive loss (3.5\% gains for R@1).
Dual Softmax further improves the results (1.6\% for R@1) and achieves an even higher state-of-the-art (37.3\% R@1).

\setlength{\tabcolsep}{3pt}
\begin{table}[t]
\begin{center}
    \resizebox{\linewidth}{!}{
    \begin{tabular}{l|cccc}
    \noalign{\smallskip}
      \textit{YouCookII}        & R@1 & R@5 & R@10 & MedR \\ 
      \midrule
    \noalign{\smallskip}
    HowTo100M \citep{miech19howto100m} &  8.2  &  24.5   &  35.3   &  24.0   \\
    \noalign{\smallskip}
    MIL-NCE \citep{miech19endtoend} &  15.1  &   38.0  & 51.2    &  10.0   \\ 
    \noalign{\smallskip}
    COOT \citep{ging2020coot} &  16.7  &   40.2  & 52.3    &  9.0   \\ 
    \noalign{\smallskip}
    UniVL \citep{Luo2020UniVL}   &   28.9  &  57.6   &   70.0   &  4.0    \\
    \midrule
    \noalign{\smallskip}
    VideoCLIP \citep{xu-etal-2021-videoclip} &  32.2  &  62.6   &   75.0   &  \bf 3.0    \\
    \noalign{\smallskip}
    \bf Ours w/o DS & 35.7 & 65.9 & 77.5 & \bf 3.0 \\
    \noalign{\smallskip}
    \bf Ours w DS &  \bf 37.6  & \bf 66.6 & \bf 78.2  & \bf 3.0 \\
    \end{tabular}}
    \caption{YouCookII Retrieval Results. DS denotes Dual Softmax.}
    \label{tab:youcook}
\end{center}
\end{table}

We provide additional results on text-video retrieval across MSR-VTT 
\footnote{We omit the results of text-video retrieval on MSR-VTT from CLIP \citep{radford2021learning} models \citep{cheng2021improving, Luo2021CLIP4Clip, fang2021clip2video, gao2021clip2tv} as it would not be a fair comparison since CLIP-based models benefit mainly from large-scale image-text pre-training, which we do not use.}
(Table \ref{tab:vtt-1k}), MSVD (Table \ref{tab:msvd}), and DiDeMo (Table \ref{tab:didemo}).
Our reported scores of VideoCLIP on MSVD and DiDeMo are from our implementation as their paper does not test on the datasets.
As \modelname\ builds on VideoCLIP \citep{xu-etal-2021-videoclip}, our results are directly comparable with the scores reported in VideoCLIP.
\footnote{
We also implemented \modelname\ in FiT \citep{Bain21}, however the improvements are not obvious as in VideoCLIP.
The reason might be the difference of video encoding in VideoCLIP and FiT.
\modelname\ contributes more to complete frame features where a video is encoded into a long sequence of video features with more temporally contextual information, rather than only a few visual frames in ViT \citep{dosovitskiy2021an} and Timesformer \citep{gberta_2021_ICML}.
}
From the additional results, it can be seen that \modelname\ outperforms VideoCLIP on all text-video retrieval datasets by a large margin.
This shows that \modelname\ is generalisable to various types of text-video retrieval data.
The smaller improvements (\eg, 30.9\% $\rightarrow$ 32.6\% R@1 on MSR-VTT 1k in Table \ref{tab:vtt-1k}) compared to those on YouCookII (32.2\% $\rightarrow$ 37.6\% R@1) might be due to the less varied scenes in shorter videos of MSR-VTT, which makes it challenging to distinguish among intra-video frames in a short video.

\begin{table}[t]
\begin{center}
\resizebox{\linewidth}{!}{
    \begin{tabular}{l|cccc}
    \noalign{\smallskip}
        \textit{MSR-VTT 1k} & R@1 & R@5 & R@10 & MedR \\ 
        \midrule
    \noalign{\smallskip}
    JSFusion \citep{Yu_2018_ECCV} &  10.2  &  31.2   &   43.2   &  13.0    \\ 
    \noalign{\smallskip}
    HowTo100M \citep{miech19howto100m} &  14.9  &  40.2   &   52.8   &  9.0    \\
    \noalign{\smallskip}
    ClipBERT \citep{lei2021less} &  22.0  &   46.8  &   59.9   & 6.0     \\ 
    \noalign{\smallskip}
    Support-set \citep{patrick2021supportset} &  30.1  &  58.5  &  69.3    &   3.0   \\ 
    \noalign{\smallskip}
    FiT  \citep{Bain21}     &  32.5   &  61.5   &  71.2    & 3.0   \\
    \midrule
    \noalign{\smallskip}
    VideoCLIP \citep{xu-etal-2021-videoclip} &  30.9  &  55.4   &   66.8   &  4.0    \\ 
    \noalign{\smallskip}
     \bf Ours    &  \underline{\bf 32.6}   & \underline{\bf 62.1}   & \underline{\bf 71.4}  & \underline{\bf 3.0}   \\
    \end{tabular}}
    \caption{MSR-VTT Results - 1k}
    \label{tab:vtt-1k}
\end{center}
\end{table}


\begin{table}[t]
\begin{center}
    \resizebox{\linewidth}{!}{
    \begin{tabular}{l|cccc}
    \noalign{\smallskip}
      \textit{MSVD}        & R@1 & R@5 & R@10 & MedR \\ 
      \midrule
    \noalign{\smallskip}
    VSE \citep{Kiros2014UnifyingVE} &  12.3 & 30.1  &  42.3  &  14.0  \\ 
    \noalign{\smallskip}
    VSE ++ \citep{faghri2018vse++} &  15.4 & 39.6  &  53.0  &  9.0  \\ 
    \noalign{\smallskip}
    CE \citep{Liu2019a} &  19.8 & 49.0  &  63.8  &  6.0  \\ 
    \noalign{\smallskip}
    Support-set \citep{patrick2021supportset} & 28.4 & 60.0  & 72.9 & 4.0   \\ 
    \noalign{\smallskip}
    FiT \citep{Bain21} & \bf 33.7 & \bf 64.7 & \bf 76.3 & \bf 3.0  \\
    \midrule
    \noalign{\smallskip}
    VideoCLIP \citep{xu-etal-2021-videoclip} &  26.4  &  52.2   &   63.3   &  5.0    \\
    \noalign{\smallskip}
    \bf Ours    & \underline{27.2}   & \underline{54.0} & \underline{64.0} &   5.0 \\
    \end{tabular}}
    \caption{MSVD Results}
    \label{tab:msvd}
\end{center}
\end{table}

\begin{table}[t]
\begin{center}
    \resizebox{\linewidth}{!}{
    \begin{tabular}{l|cccc}
    \noalign{\smallskip}
     \textit{DiDeMo}         & R@1 & R@5 & R@10 & MedR \\ 
     \midrule
    \noalign{\smallskip}
    S2VT \citep{venugopalan-etal-2015-translating} & 11.9 & 33.6  & -  & 13.0 \\ 
    \noalign{\smallskip}
    FSE \citep{DBLP:conf/eccv/ZhangHS18} & 13.9 & 36.0  & -  & 11.0 \\ 
    \noalign{\smallskip}
    CE \citep{Liu2019a} & 16.1 &  41.1 &  -  & 8.3 \\ 
    \noalign{\smallskip}
    ClipBERT \citep{lei2021less} & 20.4 & 44.5  & 56.7  & 7.0 \\
    \noalign{\smallskip}
    FiT \citep{Bain21} & \bf 31.0 & \bf 59.8  & \bf 72.4 & \bf 3.0 \\ 
    
    \midrule
    \noalign{\smallskip}
    VideoCLIP \citep{xu-etal-2021-videoclip} &  16.6  &  46.9   &   -   &  -   \\
    \noalign{\smallskip}
    \bf Ours    & \underline{19.5}   &  \underline{48.8}  &  \underline{55.9}  & 7.0   \\
    \end{tabular}}
    \caption{DiDeMo Results}
    \label{tab:didemo}
\end{center}
\end{table}

Note that video-text pairs in these downstream datasets are constructed to be aligned in order to provide strong supervision learning signals to video-language representation learning.
\modelname\ distils aligned video-text pairs and achieves noticeable improvements over approaches without any frame sampling, which corroborates our hypothesis that there are irrelevant or less useful frames in a video even if it is annotated as aligned to its text counterpart.

\begin{table}[t]
\begin{center}
\resizebox{0.7\linewidth}{!}{
    \begin{tabular}{l|c}
    \noalign{\smallskip}
     \textit{MSR-VTT QA}  & Accuracy\\ 
     \midrule
    \noalign{\smallskip}
    AMU \citep{xu2017video} &  32.5 \\
    \noalign{\smallskip}
    HME \citep{fan-CVPR-2019} &  33.0   \\ 
    \noalign{\smallskip}
    HCRN \citep{le2020hierarchical} &  35.6  \\ 
    \noalign{\smallskip}
    ClipBERT \citep{lei2021less} &  \bf 37.4    \\ 
    \midrule
    \noalign{\smallskip}
    VideoCLIP \citep{xu-etal-2021-videoclip} &  35.9   \\
    \noalign{\smallskip}
    \bf Ours & \bf 37.4 \\
    \end{tabular}}
    \caption{MSR-VTT QA Results}
    \label{tab:vttqa}
\end{center}
\end{table}

\begin{table*}[t]
\begin{center}
    \caption{Comparison of different sampling strategies for positive frames.}
    \resizebox{0.75\linewidth}{!}{
    \begin{tabular}{l|cccccc|c|ccc|c}
    \toprule
    \noalign{\smallskip}
      \textit{Strategy} & \multicolumn{6}{c|}{fixed-k ($k=1, 10, 30, 50, 100, 256$)} & \multicolumn{1}{c|}{median} &\multicolumn{3}{c|}{ratio ($30\%, 50\%, 80\%$)}  & random \\ 
      \midrule
    \noalign{\smallskip}
    R@1 & 26.90 & 30.44 & 37.17 & \bf 37.32 & 37.04 & 34.80 & \bf 37.62 & \bf 37.29 & 36.99 & 36.85 & 30.08 \\ 
    \bottomrule
    \end{tabular}
    }
    \label{tab:sampling}
\end{center}
\end{table*}

\begin{table}[t]
\begin{center}
    \resizebox{0.95\linewidth}{!}{
        \begin{tabular}{c|ccccccc}
        \toprule
        \noalign{\smallskip}
        \textit{fixed-k} & 1 & 5 & 10 & 15 & 20 & 25 & 32 \\ 
        \midrule
        \noalign{\smallskip}
        MSR-VTT QA & 35.5 & 36.3 & 36.2 & 36.8 & \bf 37.4 & 37.2 & 35.9 \\ 
        \noalign{\smallskip}
        MSR-VTT MC & 90.3 & 92.3 & 92.6 & 92.4 & 92.6 & \bf 92.7 & 92.1 \\ 
        \bottomrule
        \end{tabular}
    }
    \caption{Effect of different number of positive frames on MSR-VTT QA and MSR-VTT MC. When $k=32$, \modelname\ equals VideoCLIP.}
    \label{tab:msrvtt}
\end{center}
\end{table}

\begin{table}[h]
\begin{center}
\resizebox{0.73\linewidth}{!}{
    \begin{tabular}{l|c}
    \noalign{\smallskip}
     \textit{MSR-VTT MC}         & Accuracy\\ 
     \midrule
    \noalign{\smallskip}
    MLB \citep{kim2016hadamard} &  76.1   \\ 
    \noalign{\smallskip}
    JSFusion \citep{Yu_2018_ECCV} &  83.4 \\
    \noalign{\smallskip}
    ActBERT \citep{Zhu_2020_CVPR} &  85.7  \\ 
    \noalign{\smallskip}
    ClipBERT \citep{lei2021less} &  88.2    \\ 
    \midrule
    \noalign{\smallskip}
    VideoCLIP \citep{xu-etal-2021-videoclip} &  92.1   \\
    \noalign{\smallskip}
    \bf Ours    &  \bf 92.7  \\
    \end{tabular}}
    \caption{MSR-VTT MC Results}
    \label{tab:vttqa-mc}
\end{center}
\end{table}

\subsubsection{Video Question Answering}
Tables \ref{tab:vttqa} and \ref{tab:vttqa-mc} show the results on video question answering (VideoQA) for MSR-VTT QA and MSR-VTT MC, respectively.
For both datasets, \modelname\ improves over VideoCLIP. 
For MSR-VTT MC, it achieves a new state-of-the-art (92.7\% accuracy).
This further shows the generalisation ability of \modelname\ across different video-language tasks and datasets.

For MST-VTT QA, the score reported for VideoCLIP is from our implementation as their paper does not test on this dataset. For MSR-VTT MC, the score reported is from the original paper.
For VideoQA, we note that ClipBERT also achieves good results, which might be because it employs a multimodal Transformer encoder after two separate encoders for the video and the question to learn better cross-modal relationships.
The improvement is particularly noticeable on MSR-VTT MC, which quantitatively suggests that \modelname\ can distil question-relevant frames to improve answer accuracy.
We speculate that this is because a question only needs partial information in some frames of a video clip to be answered, which is addressed  by \modelname.

\subsection{Decision on Number of Frames}
\label{subsec:number}

Given a pair of video clip and text, we choose the positive frames according to the similarities between each frame and the text.
The number of positive frames $k$ is the key factor, deciding the set of frames to be treated as positive, and hence the extent of the contribution of the fine-grained contrastive learning signal.
We propose four strategies to choose positive frames in a video clip. 

\textbf{Fixed-k:} We select a fixed number of positive frames which have the highest similarities to the text. We experiment with $k=[1, 10, 30, 50, 100, 256]$ as the number of positive frames, with 256 as the maximum number of frames (one frame per second).\footnote{We set the maximum length of a video sequence to 256 frames for YouCookII, but 32 frames for other datasets with much shorter videos.}
\textbf{Median:} We use the averaged similarity medians in a mini-batch as the thresholds for each video: in a sequence of video frames, the ones with higher similarities than the median are used as the positive frames. The number of positive frames will vary across different mini-batches, depending on the distribution of similarities.
\textbf{Ratio:} We apply 30\%, 50\%, and 80\% of the original video length (without padding or trimming) as the positive frames. Note that different video clips have different lengths, 
so the number of sampled frames will differ from video to video.
\textbf{Random:} We randomly sample $k=50$ frames in a video clip as the positives. 
    


We show the performance of the four strategies on YouCookII in Table \ref{tab:sampling}.
\textbf{Median} has the best performance ($37.62$), which is followed by \textbf{fixed-k} with $k=50$ ($\approx 20\%$ of the data)  ($37.32$), and similarly to \textbf{ratio} with $30\%$ ($37.29$).
This indicates that on average only $\approx 20\% - 30\%$ frames in the long videos from YouCookII are informative for the retrieval task.
\textbf{Fixed-k} with $k=1$ has the lowest score, which makes sense given that the entire videos are summarised by the one most similar frame to be used as the positive candidate.
This mistakenly treats many other possibly relevant frames as negative frames, hence degrading the performance significantly.
The best number $50$ indicates that for most video-text pairs in YouCookII, 50 frames ($=$50 seconds as we extract video features at a rate of one feature per second, so the length of the extracted video features is the same as the number of seconds) ($\approx 20\%$) are the most relevant and sufficient. 
For \textbf{random}, we choose $k=50$ as this was the best number according to the fixed-k analysis.
The comparison between \textbf{random} and \textbf{fixed-k} clearly shows that sampling positive pairs based on their similarity to the text is an effective strategy to improve performance on the downstream task: on the same number of positive frames, \textbf{fixed-k} improves over \textbf{random} by 7.24\%.

We also compare the performance of \textbf{fixed-k} on MSR-VTT QA and MSR-VTT MC.
In Table \ref{tab:msrvtt}, we show that \modelname\ has the best performance on MSR-VTT QA with $k=20$ and on MSR-VTT MC with $k=25$, where both have a sequence with maximum number of $32$ frames.
The ratio of positive frames ($\approx 70\% - 80\%$) is higher than in YouCookII. This corroborates our hypothesis that fine-grained sampling is more applicable to longer videos, which tend to contain more varied scenes and where there is more scope to filter out noisy or irrelevant frames. Therefore, in video-language datasets with shorter videos,
a higher proportion of frames is needed as positive frames for effective contrastive learning.
As the number of informative frames $k$ in a video clip varies across different types of videos, we recommend that this is treated as hyperparameter that is tuned for each new dataset, following our \textbf{fixed-k} strategy to select the number $k$ on a development set. 

\subsection{Fine-grained Word Sampling}
\label{subsec:word}
Given  the improvements of \modelname\ with fine-grained frame sampling, we were curious about potential improvements if applying the same strategy to the text instead of the video, \ie\ sampling most relevant words.
Therefore, we experiment with this idea over a sequence of words to sample the most informative words as those with the highest similarity to the entire video clip in YouCookII.
The text-video retrieval results in this setup are \{R@1-32.1, R@5-62.6, R@10-75.5\}. These figures are similar to those obtained by VideoCLIP \{R@1-32.2, R@5-62.6, R@10-75.0\}, but substantially lower than our results from \modelname\ in Table ~\ref{tab:youcook}.
The reason is intuitive: by removing certain words, the meaning of the sentence or paragraph can be substantially compromised, and having an understanding of the meaning of the complete text is important for video-language tasks.
Video frames, on the other hand, can be more redundant or contribute less to the complete video understanding, and therefore fine-grained sampling from frames proves more effective. 

\subsection{Qualitative Examples}
\label{subsec:examples}
\begin{figure}[t]
    \centering
    \includegraphics[width=\linewidth]{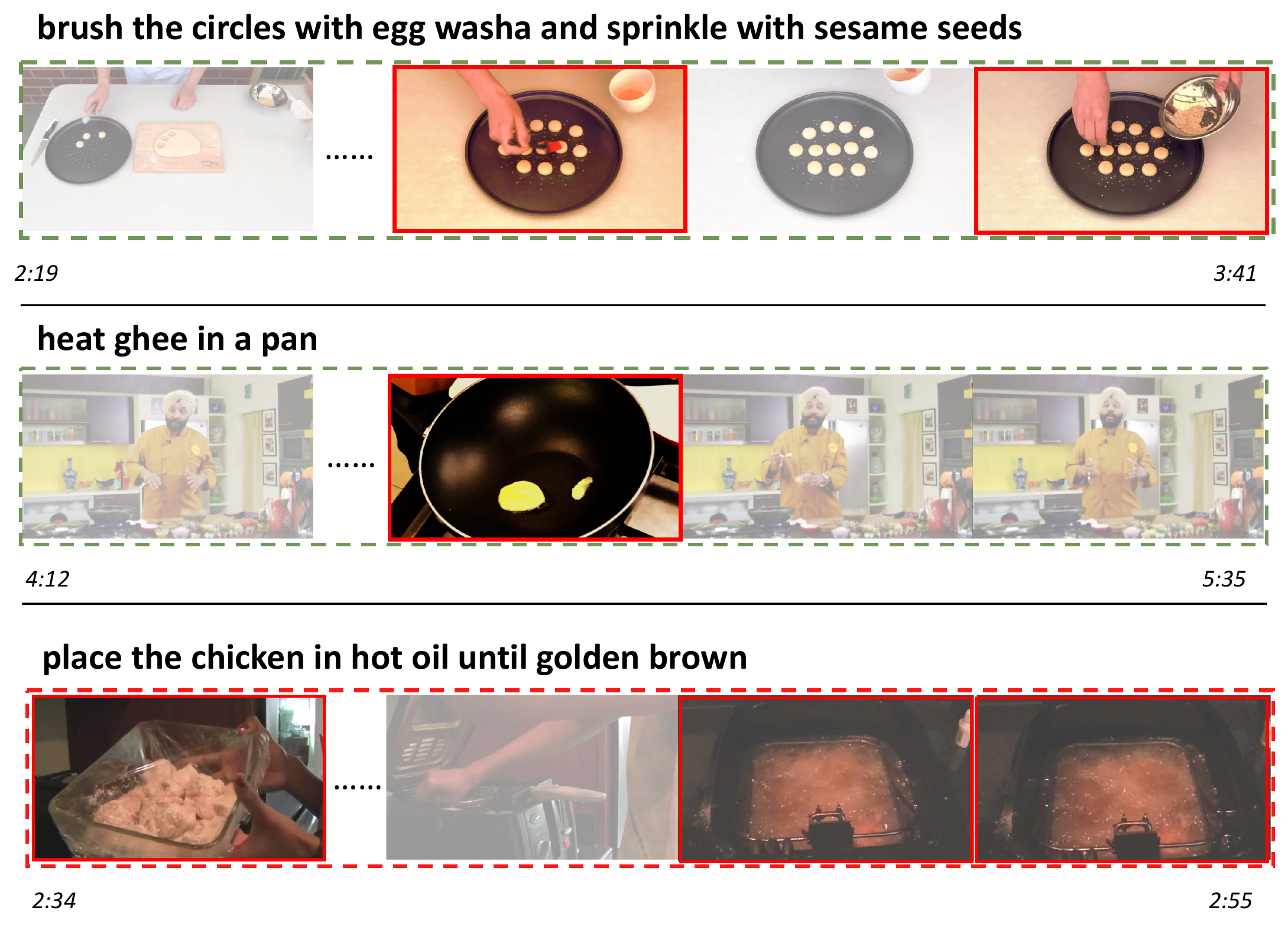}
    \caption{Qualitative examples. \modelname\ makes correct retrieval predictions on the frist two examples from YouCookII dataset. We calculate the frame-text similarities and highlight the frames with the highest scores.}
    \label{fig:examples}
\end{figure}

To further elaborate the contribution of \modelname\ and understand the effect of fine-grained contrastive loss, we show two examples where \modelname\ improves over VideoCLIP in Figure \ref{fig:examples}.\footnote{We only show a subset of informative and irrelevant frames for each example due to space limitations.}
As we can observe from the examples, some of the information in each video clip can be considered irrelevant, given the  meaning of the text. 
For example, in the first case, the long video (82 seconds) describes the cooking instruction \textit{``brush the circles with egg washa and sprinkle with sesame seeds''} but there are only two frames delivering this meaning.
This is a common feature in the YouCookII dataset, hence the positive results from sampling subsets of frames.
In the third example we show a failure case where \modelname\ does not distinguish between similar videos hence a similar but incorrect video retrieved.
We also observed failure cases where the video is either relatively short or less dynamic.  \modelname\ might not effectively distil these types of videos to find the most informative frames.
The issues could be potentially mitigated by incorporating \modelname\ into large-scale video-language pre-training to learn from more dynamic videos of various lengths.

\section{Conclusions}
We propose \modelname, an approach with a fine-grained contrastive loss to mitigate the weak correspondence problem in video-language representation learning.
Experiments conducted on text-video retrieval
and video question answering datasets 
suggest that \modelname\ can distil video frames that are relevant to its corresponding text and contribute to significant gains in performance, especially on the text-video retrieval dataset YouCookII with long videos.
\modelname\ achieves state-of-the-art on YouCookII and MSR-VTT MC, and for text-video retrieval datasets with shorter videos, it substantially improves over the base model.
Ablation studies analyse the key factors in \modelname\, including number of positive frames and word sampling. 
Our strategy for frame selection is simple and can generalise to different video-language frameworks, as long as they are based on contrastive learning, which is standard in this area. 
In addition, we posit that \modelname\ can be useful for video-language {\it pre-training} on large loosely or misaligned video-text datasets.
We hope that our work will draw attention to the need for frame-level alignment to improve video-language representation learning.

\bibliography{anthology,custom}

\end{document}